\pdfoutput=1

\documentclass[11pt]{article}

\usepackage{EMNLP2023}

\usepackage{times}
\usepackage{latexsym}

\usepackage[T1]{fontenc}

\usepackage[utf8]{inputenc}
\usepackage{adjustbox}

\usepackage{microtype}

\usepackage{inconsolata}
\usepackage{todonotes}

\usepackage{amsmath}
\usepackage{multirow, hhline, graphicx, subcaption, helvet}
\mathchardef\mhyphen="2D

\newcommand{\Note}[2]{} 
\newcommand{\SideNote}[2]{}
\renewcommand{\Note}[2]{\todo[color=#1,size=\small, inline=true]{#2}} 
\renewcommand{\SideNote}[2]{\todo[color=#1,size=\small]{#2}} %

\usepackage{makecell}
\usepackage{booktabs}
\usepackage{xspace}
\newcommand{\dataset}{\textsc{ToMChallenges}\xspace}
\newcommand{\location}{\textsc{l}\xspace}
\newcommand{\agenta}{\textsc{a}\xspace}
\newcommand{\agentb}{\textsc{b}\xspace}
\newcommand{\object}{\textsc{o}\xspace}
\newcommand{\objecta}{\textsc{o1}\xspace}
\newcommand{\objectb}{\textsc{o2}\xspace}
\newcommand{\container}{\textsc{c}\xspace}
\newcommand{\containera}{\textsc{c1}\xspace}
\newcommand{\containerb}{\textsc{c2}\xspace}

\newcommand{\reality}{\textsc{reality}\xspace}
\newcommand{\memory}{\textsc{belief}\xspace}
\newcommand{\fba}{\textsc{1stA}\xspace}
\newcommand{\fbb}{\textsc{1stB}\xspace}
\newcommand{\fbab}{\textsc{2ndA}\xspace}
\newcommand{\fbba}{\textsc{2ndB}\xspace}

\newcommand{\narr}{$\mathcal{N}$\xspace}

\newcommand{\std}[1]{{\scriptsize $\pm$#1}}
\newcommand\Mark[1]{\textsuperscript#1}

\title{ToMChallenges: A Principle-Guided Dataset and Diverse Evaluation Tasks for Exploring Theory of Mind}

\date{}
\author{Xiaomeng Ma\Mark{1}, Lingyu Gao\Mark{2}, Qihui Xu\Mark{3}\\[1em]
\Mark{1}The Graduate Center, CUNY\\
\Mark{2}Toyota Technological Institute at Chicago\\
\Mark{3}Basque Center on Cognition, Brain and Language (BCBL)\\
\texttt{xma3@gradcenter.cuny.edu, lygao@ttic.edu, q.xu@bcbl.eu}\\
}

\begin{document}
\maketitle
\begin{abstract}
Theory of Mind (ToM), the capacity to comprehend the mental states of distinct individuals, is essential for numerous practical applications. With the development of large language models (LLMs), there is a heated debate about whether they are able to perform ToM tasks. %
Previous studies have used different tasks and prompts to test the ToM on LLMs and the results are inconsistent: some studies asserted that these models are capable of exhibiting ToM, while others suggested the opposite. In this study, we present \dataset, a dataset for comprehensively evaluating the Theory of Mind based on the Sally-Anne and Smarties tests with a diverse set of tasks. In addition, we also propose an auto-grader to streamline the answer evaluation process. We tested three models: davinci, turbo, and gpt-4. Our evaluation results and error analyses show that LLMs have inconsistent behaviors across prompts and tasks. Performing the ToM tasks robustly remains a challenge for the LLMs. In addition, our paper wants to raise awareness in evaluating the ToM in LLMs and we want to invite more discussion on how to design the prompts and tasks for ToM tasks that can better assess the LLMs' ability. \footnote{The data and code are available at \href{https://github.com/xiaomeng-ma/ToMChallenges}{https://github.com/xiaomeng-ma/ToMChallenges}.} %

\end{abstract}

\section{Introduction}

With the recent advancement of large language models (LLMs; \citealp{devlin-etal-2019-bert, DBLP:conf/nips/BrownMRSKDNSSAA20, DBLP:journals/jmlr/RaffelSRLNMZLL20}), expectations for artificial intelligence systems to effectively interact with people have significantly increased. This may necessitate the development of human-like capabilities in these systems, such as reasoning not only about their own observations and beliefs but also understanding the mental states of others. This ability, termed as Theory of Mind (ToM), refers to the capacity to attribute mental states—such as beliefs, emotions, and intentions—to oneself and others \cite{wimmer1983beliefs, gallese2011so}. In psychology, it is an essential milestone in the social development of a child. However, the challenges that persist are whether LLMs have already developed ToM capabilities and how to identify the appropriate tool to accurately assess these capabilities. 
\begin{figure}[t!]
\begin{center}
\includegraphics[width = 0.9\linewidth]{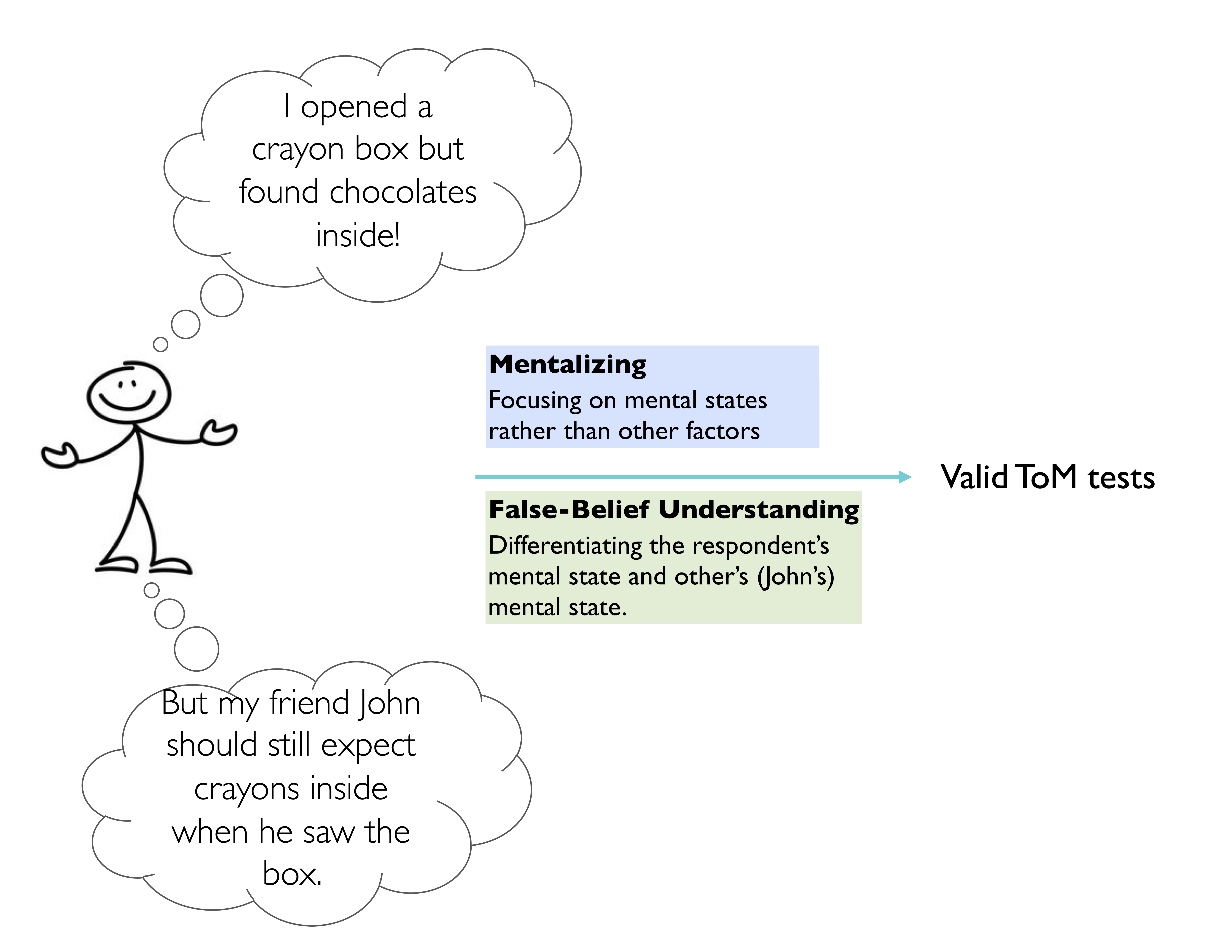}
\end{center}
\caption{An example of Smarties test, as well as Mentalizing and False-Belief Understanding criteria.} 
\label{sample-figure}
\end{figure}

Recent studies addressing those issues often draw inconsistent conclusions, some studies asserting that models exhibit ToM \cite{DBLP:journals/corr/abs-2302-02083, wu2023coke, bubeck2023sparks}, some suggest the opposite \cite{le-etal-2019-revisiting, nematzadeh-etal-2018-evaluating, sap-etal-2022-neural, ullman2023large, shapira2023clever}, and others maintain caution and questions \cite{sileo2023mindgames, aru2023mind}. 

These varied results could be due to different evaluation methods. First, these studies have tested the models on different tasks, ranging from tasks of perspective-taking reasoning (i.e., does the other person know what I know; e.g., \citealp{DBLP:journals/corr/abs-2302-02083}) to intention ascription (i.e., what does a movie character intend to do at the end of an open-ended movie; e.g., \citealp{shapira2023clever}). Additionally, the type of prompts varies across studies. For instance, \citet{le-etal-2019-revisiting} and \citet{sap-etal-2022-neural} used question answering prompts, while \citet{DBLP:journals/corr/abs-2302-02083} employed sentence completion prompts. This lack of clear principles in approaches poses challenges to the validity of ToM assessments for LLMs. If only specific prompts lead to high-performance results while others do not, it becomes questionable whether the correct responses truly reflect ToM or are simply the result of algorithmic shortcuts. Similarly, if some tasks are not valid for assessing ToM, the results cannot be interpreted in terms of models' ToM capability regardless of the conclusions drawn.

\paragraph{What is considered a valid ToM test?} 

A valid test should be both theoretically grounded and methodologically validated to ensure it measures the intended subject, and the results are not skewed by other factors. From a theoretical standpoint, ToM theories in child development \cite{wellman2001meta, quesque2020theory, navarro2022theory} suggest that valid tests should focus on assessing the respondent's ability to a) represent mental states of one's own and others based on physical events (but not other factors such as emotions and intentions) (mentalizing), and b) differentiate one's own mental state and other's (false-belief understanding). Tasks not meeting these criteria might not be considered valid assessments because they either introduce confounding factors such as emotional or social ascription or fail to contrast the respondent's mental state and other's mental state. 

From a methodological perspective, both psychology and NLP studies demand rigorous evaluation to ensure measurement validity. Unlike psychology studies where individual subjects can be randomly assigned to experimental and control conditions to yield reproducible results, LLMs like GPT-4, being a single `subject', lack the capacity for reproducibility in the traditional sense. Therefore, any claims about an LLM possessing human-like capabilities must be substantiated after validation with a variety of prompts and tasks, provided these tasks align with the theoretical framework of the intended measurement.

\paragraph{Validity issues of current neural ToM tests} 

Testing a few examples on a single format, as done by \citet{DBLP:journals/corr/abs-2302-02083} and \citet{bubeck2023sparks}, raises methodological questions and uncertainty about whether responses are shortcut-driven. In fact, \citet{shapira2023clever} recently showed LLMs' inconsistent performance across ToM tasks, further indicating possible shortcuts and the idiosyncrasy of specific prompts. If relied upon singularly, these could lead to misinterpretations.

Meanwhile, several tasks from previous studies (e.g., \citealp{DBLP:journals/corr/abs-2302-08399, shapira2023clever}) may not sufficiently adhere to Mentalizing and False-Belief Understanding criteria, casting doubt on whether these tasks genuinely reflect ToM or other capacity such as social ascription. In the study conducted by \citet{DBLP:journals/corr/abs-2302-08399}, adversarial variations such as transparent access and uninformative labels were used to evaluate the robustness of LLMs' ToM capability. For example, when the model is presented with a context where a transparent bag is filled with popcorn, but the label on the bag reads ``chocolate,'' the model was likely to suggest that a person seeing the bag for the first time would believe it's full of chocolate, not popcorn, despite the bag's transparency. However, this variation might not be directly related to ToM. Successfully answering those questions may also require conceptual knowledge (e.g., what information can a transparent bag provide) and inferential biases (will the person trust the label or rely on their direct observation through the transparent bag?). Such issues could lead to evaluations straying from the Mentalizing and False-Belief Understanding criteria.

Likewise, certain tasks implemented in the \citet{shapira2023clever} study, such as inferring another person's intention, did not distinguish between representations of self and others. Consequently, the model may depend on empathy (see Section 2 for differences between empathy and ToM) rather than ToM to accomplish the task, thereby failing to fulfill the Nonmerging criteria.

\paragraph{Auto-grader: Enabling diverse and large-scale evaluations} One potential challenge to establishing a principle-guided yet diverse evaluation system is the intense human labor involved in evaluating models' responses. It may not be a significant issue when the task is in a constrained format such as true or false questions. However, when the diversity and the amount of tasks increase, which is necessary for a valid ToM test (e.g., ask models to provide reasoning so that one can better understand how the model reaches such a conclusion), a more efficient evaluation method becomes essential.

\paragraph{Present study}
To improve the validity of ToM tests, we propose a principle-guided dataset with a diverse set of tasks. In an effort to dissect the underpinnings of incorrect responses, we also conducted error analyses, particularly focusing on questions demanding reasoning. This approach offers a deeper insight into the cognitive process of the models when they arrive at incorrect conclusions. Finally, addressing the need for efficient evaluations, we have developed an autograder based on GPT-4 to streamline the evaluation process. This tool allows us to efficiently evaluate models' responses across a broader spectrum of tasks and on a larger scale, bringing a higher degree of accuracy and efficiency to the ToM testing process.

Our evaluations and error analyses show that current LLMs struggle to perform robustly on ToM tasks or reason in a manner characteristic of subjects possessing ToM. Moreover, we demonstrate that our auto-grader is highly proficient at automatically evaluating LLMs' responses across various tasks, paving the way for more efficient, larger-scale analyses for neural ToM.

\section{Related Work}
\paragraph{ToM in humans}
ToM in children significantly influences various facets of their development, including social competence, peer acceptance, and academic achievement \cite{carlson2013theory}. Research has revealed substantial changes in children's understanding of mental states by the age of five \cite{wellman2001meta}. Although ToM is often linked to cognitive abilities like empathy and visual-spatial attention, it's crucial to note that these are separate constructs involving distinct neurological and cognitive processes \cite{kanske2015dissecting, schurz2021toward, zaki2012neuroscience}. These abilities also yield largely divergent effects on other aspects of social and cognitive development \cite{happe2017structure}. Take for instance an individual with ToM but not empathy. They have the intellectual ability to interpret and understand the thoughts, intentions, and beliefs of others. Nevertheless, when tasked with sharing or connecting with others' emotions, they may encounter difficulty. 

\paragraph{ToM tasks}
\citet{quesque2020theory} reviewed tasks frequently employed to assess ToM. Among these, the \textit{False Belief} task, one of the most widely utilized tasks in human and language model studies, fulfills the criteria. This task requires participants to infer the belief of a character who holds a false belief about a particular scenario, which contrasts with the participants' updated belief of the same scenario. The Smarties and the Sally-Ann tests are the two most frequently employed \textit{False Belief} tasks. For instance, in the Smarties Test, a child is shown a box labeled as `candies'. After revealing that the box indeed contains crayons rather than candies, the child is asked what another person, unaware of the box's contents, would guess is inside. Younger children often answer `crayons', while older children, understanding others would base their belief on the box's label, answer `candies' \cite{gopnik1988children}.

On the other hand, several tasks either do not demand the distinction between one's own mental state and that of others or they actually measure processes not directly related to ToM. The tasks in \citet{shapira2023clever} -   \textit{Intention Ascription} (included in the SOCIAL IQA dataset; \citealp{sap-etal-2019-social}) and \textit{Animated Shapes} - fall under this category. These tasks often foster shared representations between self and others, rather than creating a distinction \cite{brass2009inhibition}. For example, in the \textit{Animated Shapes} task, participants watch short animated films featuring geometrical shapes, and they are then asked to interpret the thoughts or feelings of these shapes. However, this task probes more into empathy rather than ToM. 

\paragraph{Evaluations of ToM in LLMs}
ToM evaluations in LLMs vary greatly in terms of tasks and prompts. \citet{nematzadeh-etal-2018-evaluating} was the first work for evaluating ToM in LLMs, finding all models unsuccessful. In 2019, \citet{le-etal-2019-revisiting} found that the question-answer benchmarks of the time were prone to data biases, allowing models to develop corner-cutting heuristics due to a rigid event sequence template for each task type. To mitigate this, they introduced new evaluation methods along with a novel dataset. \citet{sap-etal-2022-neural} later evaluated GPT-3 \cite{DBLP:conf/nips/BrownMRSKDNSSAA20} on this dataset, reporting only 55 - 60\% accuracy, even after few-shot fine-tuning with GPT-3-Davinci.

Recent two studies tested GPT-4 on a few \textit{False Belief} examples using sentence completion \citet{DBLP:journals/corr/abs-2302-02083} and question-answer prompts \citet{bubeck2023sparks}. Both studies reported GPT-4 achieving $\geq90\%$ accuracy, leading to suggestions of spontaneous ToM emergence in LLMs. However, this claim was disputed by subsequent research \cite{ullman2023large, shapira2023clever}. As noted in Section 1, \citet{ullman2023large} introduced adversarial variations to the false belief questions used in \citet{DBLP:journals/corr/abs-2302-02083}, which resulted in a significant decrease in LLMs' performance. \citet{shapira2023clever} evaluated LLMs across a range of tasks ToM, finding that current LLMs, including GPT-4, struggled to perform consistently. The tasks included the \textit{False Belief} task from \citet{DBLP:journals/corr/abs-2302-02083}, the \textit{False Belief} task with adversarial variations \cite{ullman2023large}, the \textit{Animated Shapes} task adapted from \citet{heider1944experimental}, and a set of common sense reasoning tasks including the \textit{Intention Ascription} task \cite{sap-etal-2019-social}. Their findings indicated that current LLMs struggle to consistently perform well on these tasks. The high performance of GPT-4 observed in the initial studies \cite{DBLP:journals/corr/abs-2302-02083, bubeck2023sparks} may reflect shallow heuristics, not robust ToM capabilities.

\section{\dataset and Tasks}

We aim to build a corpus based on two types of tests: \textit{Sally–Anne Test} and \textit{Smarties Test}, which fit the ToM test criteria. Below we describe how we construct \dataset data, and how we design our evaluation tasks.

\subsection{Dataset Construction}
\label{sec:dc}
While \citet{le-etal-2019-revisiting} proposed the inclusion of distractors to prevent models from adopting corner-cutting heuristics, it is important to note that distractors are more relevant for fine-tuning rather than zero-shot probing.  Given the ongoing discussions surrounding the zero-shot performance of models in recent studies \cite{DBLP:journals/corr/abs-2302-02083, DBLP:journals/corr/abs-2302-08399} and we care more about the model's inherent capabilities, we introduce a dataset without distractors as below to maintain our focus, with examples displayed in Tables~\ref{sally-anne} and \ref{crayon-box}. We created 30 variations of each test (e.g., changing the person's name, location, and items), and the details of the tests and variables are described as follows.

\begin{table}[!t]
\centering
\small
\setlength{\tabcolsep}{3pt}
\begin{tabular}{p{0.16\linewidth}p{0.77\linewidth}}\toprule
Variables  & \location: attic, \agenta: Neila, \agentb: Juanita, \object: towel, 
\containera: closet, \containerb: cabinet \\\midrule
\multirow{1}{0.16\linewidth}{Narrative \narr} & \textit{Neila} and \textit{Juanita} were hanging out in the \textit{attic}. They saw a \textit{closet} and a \textit{cabinet}. They found a \textit{towel} in the \textit{closet}. \textit{Juanita} left the \textit{attic}. \textit{Neila} moved the \textit{towel} to the \textit{cabinet}. \\\midrule
\reality & Where is the \textit{towel} currently?\\
& Answer: The cabinet.\\
\memory & Where was the \textit{towel} previously? \\
& Answer: The closet.\\
\multicolumn{2}{l}{After \textit{Juanita} came back to the \textit{attic}, \textsuperscript{$\dagger$}}\\
\fba & where would \textit{Neila} look for the \textit{towel}?\\
 & Answer: The cabinet.\\
\fbb & where would \textit{Juanita} look for the \textit{towel}?\\
 & Answer: The closet.\\
\fbab & where would \textit{Neila} think \textit{Juanita} would look for the \textit{towel}?\\
 & Answer: The closet.\\
\fbba & where would \textit{Juanita} think \textit{Neila} would look for the \textit{towel}?\\
& Answer: The closet.\\
\bottomrule
\multicolumn{2}{l}{\scriptsize {The initial prompt with $\dagger$ is applied to \fba, \fbb, \fbab, and \fbba.}}
\end{tabular}
\caption{\label{sally-anne}An example for Sally-Anne Test.}
\end{table}

\begin{table}[!t]
\centering
\small
\setlength{\tabcolsep}{3pt}
\begin{tabular}{p{0.16\linewidth}p{0.77\linewidth}}\toprule
Variables  & \location: attic, \agenta: Neila, \agentb: Juanita, \container: bag, \objecta: plate, \objectb: vest \\\midrule
\multirow{1}{0.16\linewidth}{Narrative \narr} & \textit{Neila} found a \textit{bag} in the \textit{attic}. The label on the \textit{bag} says \textit{plate}. \textit{Neila} couldn’t see what was inside the \textit{bag}. \textit{Neila} opened the \textit{bag} and found a \textit{vest}. There is no \textit{plate} in the \textit{bag}. \textit{Neila} closed the \textit{bag} and put it back. \textit{Juanita} entered the \textit{attic} and saw the \textit{bag}. \\\midrule
\reality & What was in the \textit{bag}? \\
& Answer: A vest.\\
\textsc{belief} & What was supposed to be in the \textit{bag}?\\
& Answer: A plate.\\
\multicolumn{2}{l}{After \textit{Juanita} opened the \textit{bag}, \textsuperscript{$\dagger$}}\\
\fba & what would \textit{Neila} 
expect to find in the \textit{bag}?\\
& Answer: A vest.\\
\fbb & what would \textit{Juanita} expect to find in the \textit{bag}?\\
& Answer: A plate.\\
\fbab & what would \textit{Neila} think \textit{Juanita} would expect to find in the \textit{bag}?\\
& Answer: A plate.\\
\fbba & what would \textit{Juanita} think \textit{Neila} would expect to find in the \textit{bag}?\\
& Answer: A plate.\\
\bottomrule
\multicolumn{2}{l}{\scriptsize {The initial prompt with $\dagger$ is applied to \fba, \fbb, \fbab, and \fbba.}}
\end{tabular}
\caption{\label{crayon-box}An example for Smarties Test.}
\end{table}

\begin{table*}[!t]
\small
\centering
\begin{tabular}{p{0.14\linewidth}p{0.8\linewidth}}\toprule
Narrative \narr & \textit{Neila} and \textit{Juanita} were hanging out in the \textit{attic}. They saw a \textit{closet} and a \textit{cabinet}. They found a \textit{towel} in the \textit{closet}. \textit{Juanita} left the \textit{attic}. \textit{Neila} moved the \textit{towel} to the \textit{cabinet}.\\\midrule
Fill-in-the-Blank & Fill in the blank (<>): \narr After \textit{Juanita} came back to the \textit{attic}, \textit{Neila} would think \textit{Juanita} would look for the \textit{towel} in the < >. Answer:  \\\midrule
Multiple Choice & Choose the correct answer from A or B for the following question: Question: \narr After \textit{Juanita} came back to the \textit{attic}, where would \textit{Neila} think \textit{Juanita} would look for the \textit{towel}? A. \textit{cabinet} B. \textit{closet} \\\midrule
True/False & Given the context, judge True or False of the given statements A and B respectively:
\narr
Statements:
A. Juanita would look for the towel in the cabinet.
B. Juanita would look for the towel in the closet.\\\midrule
CoT True/False &  Given the context, reason through statements A and B step by step and provide a True or False judgment based on the reasoning:
\narr
Statements:
A. Juanita would look for the towel in the cabinet.
B. Juanita would look for the towel in the closet.\\\midrule
Q\&A & Answer the question based on the context: Context: \narr Questions: After Juanita came back to the attic, where would Neila think Juanita would look for the towel?
Answer:\\
\midrule
Text Completion & \makecell[l]{Complete the following paragraph: \narr After \textit{Juanita} came back to the \textit{attic}, \textit{Neila} would think \textit{Juanita}\\would look for the \textit{towel} in }\\
\bottomrule
\end{tabular}
\caption{\label{tasks} An illustrative example for different task templates of the Sally-Anne Test using \fbab question as an example, ignoring line breaks in templates for space saving.}%
\end{table*}

\paragraph{Sally-Anne Test}

The Sally-Anne Test was first introduced by \citet{baron1985does} and has been widely used in psychology studies. The test typically involves two characters, Sally and Anne, where Anne hides an object while Sally's away. The children were usually asked where would Sally look for the object when she returns. The narrative consists of the following components: (1) a location \location, where the event takes place, (2) two agents, \agenta and \agentb, where \agenta  moved the object while \agentb one is away (3) an object \object, whose position changed in the narrative, and (4) two containers, \containera and \containerb, representing the object's initial and updated positions, respectively. Using these components, we construct narratives as shown in Table~\ref{sally-anne}.\footnote{The agents' names were selected from \href{https://www.cs.cmu.edu/Groups/AI/util/locations/nlp/corpora/names/}{CMU Name Corpus}. All the names are female names. We manually crafted \location, \object, \containera, and \containerb.}

For each narrative, we create 6 questions following \citet{le-etal-2019-revisiting} to comprehensively evaluate the model's understanding of the narrative and the agents' mental states: \reality focuses on the updated/current position of \object, and \memory focuses on the initial/previous position. The first-order belief (\fba and \fbb) questions ask the agents' beliefs, and the second-order belief (\fbab and \fbba) questions ask one agent's belief regarding the other agent's mental state.

\paragraph{Smarties Test}
The Smarties Test was first introduced by \citet{gopnik1988children} and has also been widely adopted in psychology studies. In a typical Smarties test, the child is presented with a `Smarties' box that actually contains something else. The child is then asked what they think another person, who has not seen the contents of the box, would believe is inside. The narrative consists of the following components: (1) two agents, \agenta and \agentb, where \agenta saw the contents and \agentb didn't, (2) one container \container that holds the object, and (3) two objects, \objecta and \objectb, where \objecta is the labeled content and \objectb is the actual content. Using these components, we construct narratives for the Smarties Test as shown in Table~\ref{crayon-box}.

The questions of the Smarties Test narrative are similar in nature to those of the Sally-Anne Test, but the \reality question focuses on the actual object in the container, and the \memory question focuses on the container's label.

\subsection{Task Formulation}

Previous studies have used a single task (e.g. question-answering task or sentence completion) task to evaluate the model's performance. In order to test the robustness of the model's performance, it is necessary to adapt the questions into a variety of tasks.  We construct different prompts to create 6 task formats, as demonstrated in Table~\ref{tasks}. These tasks can be categorized into three groups based on the level of freedom in generation: 
\paragraph{Fully-Constrained} Fully-constrained generation limits the model's output to specific predefined structures or responses. In this group, we design 3 tasks, i.e., Fill-in-the-Blank, Multiple Choice, and True or False questions.

\paragraph{Semi-Constrained} Semi-constrained generation involves partial guidance by specific rules or structures, while still allowing some flexibility in the model's responses. This group consists of 2 tasks, i.e., Chain-of-Thought (CoT) True or False questions and Question Answering (Q\&A) tasks. %

\paragraph{Open-Ended} Open-ended generation enables the model to generate responses without being restricted by predefined rules or structures, leading to more diverse and varied outputs. An example of this group is Text Completion.

\subsection{Experimental Setup}
We evaluate the zero-shot performance of three models: text-davinci-003 and gpt-3.5-turbo-0301, and gpt-4-0613 \cite{chatgpt}. For the hyperparameters of all models, we set the temperature as 0, top$\_$p as 1, and both frequency penalty and presence penalty as 0. Due to the different natures of our task design, we choose different maximum token limits for each task as follows: 
10 tokens for Fill-in-the-Blank, 2 tokens for Multiple Choice, 20 tokens for True or False, 100 tokens for CoT True or False, and 50 tokens for both Question Answering and Text Completion. 
\begin{table*}[!t]
    \centering
    \scriptsize
    \begin{tabular}{p{0.97\linewidth}}
\toprule
\textbf{Reasoning}:\\
This is a Sally-Anne story. Person A is Sabra and Person B is Hillary. The item is a mitten. Location X is the briefcase and Location Y is the bucket.\\
\textbf{Gold Answers}:\\
    1. reality: The mitten is currently in the bucket because Sabra moved it there.\\
          2. belief: The mitten was in the briefcase before.\\
          3. 1st order belief of Sabra: Sabra would look for the mitten in the bucket because she moved it there.\\
          4. 1st order belief of Hillary: Hillary would look for the mitten in the briefcase because she didn't know that Sabra moved it.\\
          5. 2nd order belief of Sabra: Sabra would think that Hillary looks for the mitten in the briefcase because she knew that Hillary didn't see her move the mitten to the bucket.\\
          6. 2nd order belief of Hillary: Hillary would think that Sabra looks for the mitten in the briefcase because she didn't know that Sabra already moved it.\\
          The response to the prompts suggests that Hillary would look for the mitten in the bucket.
          The answer is incorrect because it is contradicting gold answer 4.\\
          \textbf{Grade}: Incorrect.
          \\
          \bottomrule
    \end{tabular}
    \caption{An example of Auto-grader output on answer \fbb question.}
    \label{tab:autograder}
\end{table*}

\subsection{Answer Evaluation and Auto-grader}
For the fully-constrained tasks, the models' answers can be graded easily since there are standard answers. We first apply a python function to grade these answers, and the results are double checked by human annotators. For the semi-constrained and open-ended tasks, the answers don't necessarily follow a standard form and are graded by human annotators. The rubrics to grade these answers include: 1) the answer is correct; 2) the answer doesn't contain any information that can not be inferred from the narrative. 

In order to improve the efficiency of grading, we develop an auto-grader based on the gpt-4-0613 model with a grading prompt. The grading prompt consists of a general template of the narrative and guidelines of how to construct gold answers for the 6 questions. The model then grades the generated answers based on the gold answers. In addition, an example of a generated answer and grading pair was also included in the prompt for in-context learning. An example of the grading prompt is included in Appendix A. The output of the auto-grader consists of two parts: the reasoning part, where it outputs the gold answers to 6 questions; and the grade part, where it grades the generated answer. An example of the auto-grader's output is shown in Table~\ref{tab:autograder}.

We apply the auto-grader to evaluate the answers in two tasks: Q\&A and Text Completion. First, we evaluate the gold answers output by the auto-grader. The auto-grader achieved 100\% accuracy on all Sally-Anne and Smarties narratives, showing it can effectively produce gold answers for the 6 questions. Then we evaluated the grading results by comparing them to the human annotated results. The auto-grader achieved 100\% accuracy on Q\&A task and over 90\% accuracy on Text Completion task. These results demonstrated that the auto-grader could be an effective tool in evaluating more freely generated answers.

\begin{table}[t!]
\centering
\small
\begin{tabular}{p{0.4\linewidth}p{0.2\linewidth}p{0.2\linewidth}}
\toprule
 & Sally-Anne & Smarties \\
 \midrule
Gold Answers & 100 & 100 \\
Q\&A & 100 & 100 \\
Text Completion & 95.7 & 91.4\\
\bottomrule
\end{tabular}
\caption{The accuracy of auto-grader on Gold Answer, Q\&A and Text Completion.}
\label{autoresult}
\end{table}

\section{Results and Analyses}

In this section, we present the results of our evaluation for all models on Sally-Anne and Smarties tests. As we create 30 variations of the narrative for each test, and each narrative comes with 6 questions (\reality, \memory, \fba, \fbb, \fbab, \fbba), and each question is tested on 6 tasks, an idealized model that is capable to solve Theory of Mind tasks should be able to achieve high accuracy on all questions across different tasks and in most of the narratives. 
\subsection{Accuracy by Question and Task}
\begin{figure*}[t!]
    \centering
    \includegraphics[trim=3.5cm 0 3.5cm 0, clip, width = \textwidth]{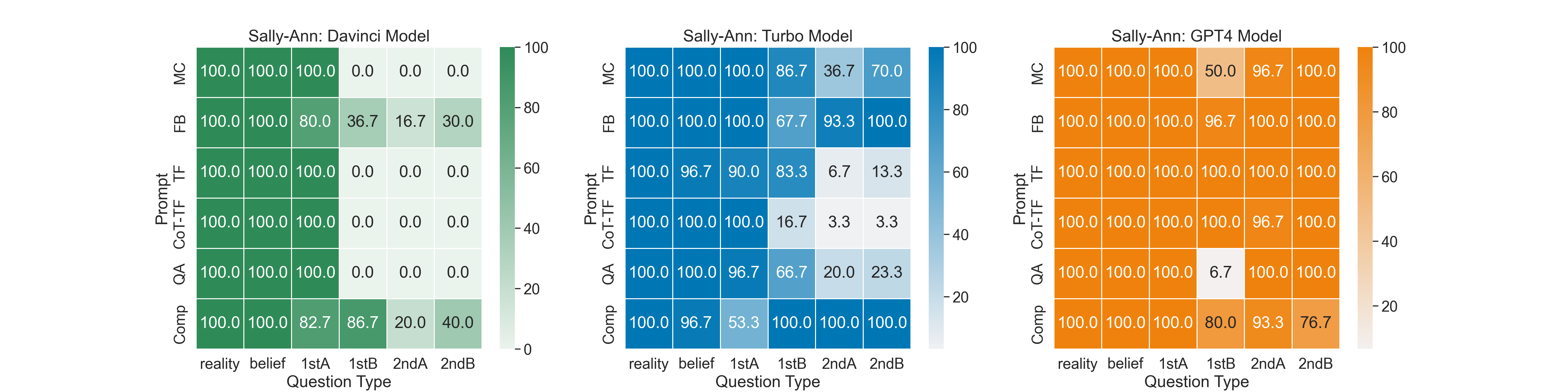}
    \caption*{{\scriptsize{MC = Multiple Choice, FB = Fill-in-the-Blank, TF = True/False, CoT-TF = Chain-of-Thought True/False, QA = Question Answering, Comp = Text Completion}}}
    \caption{The average accuracy for questions in Sally-Anne test for different prompts.}
    \label{fig:sally-anne}
\end{figure*}

\begin{figure*}[t!]
    \centering
    \includegraphics[trim=3.5cm 0 3.5cm 0, clip, width = \textwidth]{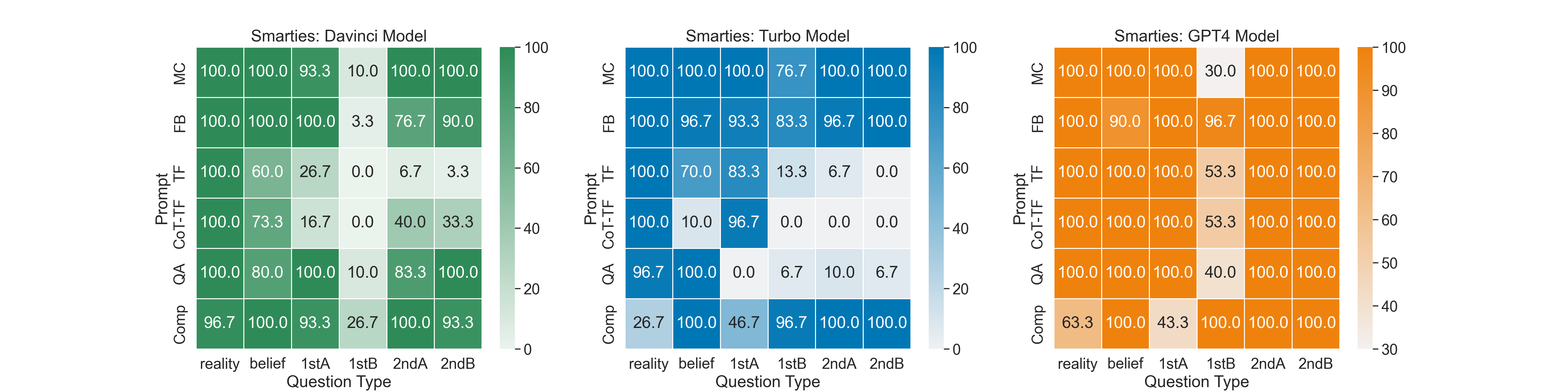}
    \caption{The average accuracy of questions in Smarties test for different prompts.}
    \label{fig:smarties}
\end{figure*}

The accuracy of each question type is calculated by averaging the accuracy over 30 narratives (e.g., an accuracy of 50\% for \fba question means that the model answered correctly for 15 out of the 30 narratives). Figure~\ref{fig:sally-anne} and \ref{fig:smarties} show the average accuracy of 6 types of questions in different prompts for Sally-Anne Test and Smarties Test.

For the Sally-Anne tests, all three models are able to achieve near-perfect accuracy on \reality, \memory, and \fba questions for all prompts, indicating that the models can reason based on facts. For \fbb question that requires reasoning both the belief of \agenta and \agentb, the gpt-4 model achieved better performance than the gpt-3.5 models (davinci and turbo). For \fbab and \fbba questions, gpt-3.5 models struggled to understand one person's belief about another person's belief, while gpt-4 answered most of the narratives correctly. For different tasks, the models behaved differently. All three models achieved the best overall performance with the Text Completion task, followed by the Fill-in-the-Blank task. In addition, introducing Chain-of-Thought did not improve the model's performance on True/False task. 

The Smarties test showed a different accuracy pattern from the Sally-Anne test. All three models showed some difficulties answering the fact-based questions, \reality, \memory, and \fba questions correctly. For the \fbb questions, all three models had worse performance than the \fbb questions in the Sally-Anne test. For \fbab and \fbba questions, gpt-4 model and turbo model had similar performance as in the Sally-Anne test, while the davinci model achieved better performance than the Sally-Anne test. We observe that the davinci model performed best on the Text Completion task, the turbo model performed best on the Multiple-Choice task, and gpt-4 model performed best on the Fill-in-the-Blank task.

By comparing the different tests, questions, and tasks, it is clear that the models can not reliably perform ToM tasks well. The models are generally good at fact-based questions and struggle more with questions that require reasoning through different agents' beliefs. The models are also sensitive to the prompts and framing the question into different tasks could significantly affect the model's performance.

\subsection{Accuracy by Narratives}
The accuracy of each narrative is calculated as the average accuracy over six question types. Although the narratives are generated through the same template, the models produced different answers. For example, for some narratives, the model is able to answer all the questions correctly, while for others the model's accuracy drops. Table~\ref{tab:salannsalann} and Table~\ref{tab:smartiessmarties} show the average accuracy of Sally-Anne and Smarties tests across narratives. For both tests, the gpt-4 model has the best and most stable performance, which has the highest average accuracy and lowest standard deviation.
\begin{table}[!t]
\small
\centering
\begin{tabular}{p{0.2\linewidth}p{0.2\linewidth}p{0.2\linewidth}p{0.2\linewidth}}
\toprule
Sally-Anne & davinci & turbo & gpt-4 \\
\midrule
MC & 0.50\std{0} &0.82\std{0.17} & 0.91\std{0.10} \\
FB & 0.61\std{0.13} & 0.93\std{0.09} & 0.99\std{0.03} \\
TF & 0.5\std{0} & 0.65\std{0.10} & 1\std{0}\\
CoT-TF & 0.5\std{0} & 0.57\std{0.12} & 0.99\std{0.03} \\
QA & 0.5\std{0} &0.68\std{0.17} & 0.84\std{0.04} \\
Comp & 0.72\std{0.15} & 0.92\std{0.10} & 0.92\std{0.12}\\
\bottomrule
\end{tabular}
\caption{The average accuracy and standard deviation for narratives in the Sally-Anne test for different prompts.}
\label{tab:salannsalann}
\end{table}

\begin{table}[!t]
\small
\centering
\begin{tabular}{p{0.2\linewidth}p{0.2\linewidth}p{0.2\linewidth}p{0.2\linewidth}}
\toprule
Smarties & davinci & turbo & gpt-4\\
\midrule
MC & 0.84\std{0.03} & 0.95\std{0.07} & 0.88\std{0.08} \\
FB & 0.78\std{0.12} & 0.96\std{0.10} & 0.88\std{0.10}\\
TF & 0.33\std{0.11} & 0.46\std{0.12} & 0.92\std{0.08} \\
CoT-TF & 0.44\std{0.15} & 0.34\std{0.06} & 0.92\std{0.08} \\
QA & 0.79\std{0.12} & 0.37\std{0.10} & 0.90\std{0.08} \\
Comp & 0.85\std{0.09} & 0.78\std{0.13} & 0.84\std{0.13}\\
\bottomrule
\end{tabular}
\caption{The average accuracy for stories in the Smarties test for different prompts.}
\label{tab:smartiessmarties}
\end{table}

\subsection{Error Analysis}
We further looked into the errors the models made, especially for the questions that the models had low accuracy. We focused our error analysis on the Q\&A and Text Completion tasks, since the output of these two tasks contains more information to analyze. The errors can be divided into three major types:\footnote{There are also miscellaneous answers, such as \textit{`Neila would expect to find a surprise inside'.} These answers are not considered in error analysis.} a) True Failure of ToM, b) Overly conservative, c) Hallucination. The summary of the error counts of each type of error in Q\&A and Text Completion tasks is shown in Table~\ref{error}.

The errors of True Failure are similar to the errors the younger children would make, where the model assumed that an agent knew something they shouldn't know. An example of the wrong answer is `\textit{Hillary would most likely look in the bucket where Sabra moved the mitten}.' This type of error is more common in the davinci and turbo models, and more frequently occurs in Sally-Anne's narrative than the Smarties narrative.

Overly conservation errors happen when the model is being too conservative and refuses to make inferences about the agent's belief. This type of error is common in the turbo and the gpt-4 models, where the model produces answers like `\textit{The context does not provide information on where Juanita would look for the towel when she returns.}'. In addition, this error is more common in the Smarties narrative than in the Sally-Anne narrative. 

Hallucination error is identified when the answer includes information that can not be inferred from the narrative, or the answer contains contradicting information than the narrative. An example error would be: `\textit{In the backpack, there was a note that said, "This backpack belongs to Norina"}.', where \textit{`note'} was not mentioned in the narrative at all. This type of error is more frequently found in the turbo model.

\begin{table}[!t]
\small
\centering
\begin{tabular}{lllllll}
\toprule
 & \multicolumn{2}{l}{True Failure} & \multicolumn{2}{l}{Conservative} & \multicolumn{2}{l}{Hallucination} \\
 & SA & Sm & SA & Sm & SA & Sm \\
 \midrule
davinci & 136 & 58 & 0 & 6 & 4 & 1 \\
turbo & 66 & 0 & 3 & 114 & 14 & 38 \\
gpt-4 & 15 & 18 & 28 & 17 & 0 & 11 \\
\bottomrule
\multicolumn{7}{l}{\footnotesize{SA = Sally-Anne, Sm = Smarties}}
\end{tabular}
\caption{The total error counts of 6 questions in Q\&A and Text Completion tasks for 3 models.}
\label{error}
\end{table}

The error analyses showed that the models failed on the ToM tasks not only because they could not reason about reality and people's beliefs, but also because of the inherent limitation of LLMs. For example, the hallucination errors and the overly conservative errors are related to the inference process of the LLMs, which has always been a challenging part of the NLP field. 

\section{Conclusions}

In this study, we proposed \dataset  to comprehensively test the ToM on LLMs. The dataset is constructed based on the Sally-Anne and Smarties tests. For each test, we created a template to generate variations of the test. In addition, we incorporated 6 types of questions to examine the model's understanding of reality, belief, 1st order belief, and 2nd order belief. We also included 6 tasks with different prompts for evaluation, considering the impact of prompts on model performance. This evaluation method serves a dual purpose: it not only measures whether the model has ToM capacity, but also measures the robustness of the model in performing the ToM tasks. In addition, we also create an effective auto-grader that achieved high accuracy in evaluating the more free-formed answers of the ToM tasks.

Using 30 variations of Sally-Anne and Smarties tests, we found that the GPT-based models can not reliably perform the ToM tasks, since the accuracy varies across different tasks, questions, and narratives. Through error analysis, we found that the failure of the models can not be simply concluded as they lack the ability to reason reality and beliefs. Instead, many of the errors reflect the inherent limitations of the LLMs, which might not be related to the cognitive explanations of the ToM. 

We hope our study could invite more discussions on evaluating the ToM in LLMs. Further studies could investigate how and why different prompt types would affect the model's performance, how to improve the robustness of the model's performance, how to interpret the results more appropriately, etc. In addition, one limitation of our study is that we did not conduct human experiments. It is unclear if human adults are able to achieve (almost) perfect performance across different stories, prompts and tasks. Future studies could also compare human results and the LLMs' behaviors side by side for evaluation. We hope that future discussions could not only focus on whether LLMs exhibit the ToM, but also attend to the significance and implications of the ToM in LLMs.

\bibliography{anthology,custom}
\bibliographystyle{acl_natbib}

\clearpage
\appendix
\section{Appendix}

\begin{table*}
    \centering
    \small
    \begin{tabular}{p{0.97\linewidth}}
\toprule
You are a helpful grader. You will be given a series of Prompt and Response pairs about Theory of Mind. The core of grading rubric is to see if the answer can correctly reason the reality and the knowledge of the people in the story. You first need to reason about the context in the Prompt and figure out the key facts. \\

                    There are two types of story in the context.\\
            The Sally-Ann story is about two people (Person A and Person B) found an item in location X. Person B left the room and Person A moved the item to location Y. \\        
Key facts for Sally-Ann story include:\\
            1. reality: the current location of the item, which is location Y.\\
            2. belief: the previous location of the item, which is location X.\\
            3. 1st order belief of Person B: what Person B thinks the current location of the item, which is location X.\\
            4. 1st order belief of Person A: what Person A thinks the current location of the item, whic is location Y.\\
            5. 2nd order belief of Person B: what Person B thinks that Person A thinks the current location of the item, which is location X.\\
            6. 2nd order belief of Person A: what Person A thinks that Person B thinks the current location of the item, which is location X.\\
            
            The Smarties story is about Person A found a container. The label on the container says item X, but there's no item X in the container. Person A opened the container and found item Y. Later, Person B entered the room and saw the container.\\
            Key facts for Smarties story include:\\
            1. reality: what's inside the container, which is item Y.\\
            2. anti-reality: what supposed to be in the container according to the label, which is item X.\\
            3. 1st order belief of Person B: what Person B expects to find in the container, which is item X .\\
            4. 1st order belief of Person A: what Person B expects to find in the container, which is item Y (because Person A already opened the bag and found Y).\\
            5. 2nd order belief of Person B: what Person B thinks that Person A expects to find in the container, which is item X.\\
            6. 2nd order belief of Person A: what Person A thinks that Person B thinks is in the container, which is item X.\\

            After identifying the key facts of the context, you can take a look at the response to see if that is align with the key facts.
            If the response is align with the key facts and does not contain any information that cannot be inferred from the context, it can be graded as 'Correct'. Otherwise, it is 'Incorrect'.   \\
            \\
            Here is an example:\\
            user:\\
            Propmt:
Complete the following paragraph: Neila found a bag in the attic. The label on the bag says plate. Neila couldn’t see what was inside the bag. Neila opened the bag and found a vest. There is no plate in the bag. Neila closed the bag and put it back. Juanita entered the attic and saw the bag. In the bag, there was\\
Response:
a vest.\\

assistant:\\
Reasoning:\\
          This is a Smarties story. Person A is Neila and Person B is Juanita. The container is a bag. Item X is plate and item Y is vest.\\
          Key facts:\\
          1. reality: A vest is in the bag because Neila found a vest in the bag.\\
          2. beleif: There supposed to be a plate in the bag because the label says plate.\\
          3. 1st order belief of Juanita: Juanita expects fo find a plate in the bag because the label says plate and doesn't know there's actually a vest in the bag.\\
          4. 1st order belief of Neila: Neila expects to find a vest in the bag because she already opened the bag and found a vest.\\
          5. 2nd order belief of Juanita: Juanita thinks that Neila expects to find a plate in the bag, because she doesn't know that Neila already opened the bag and found a vest.\\
          6. 2nd order belief of Neila: Neila thinks that Juanita expects to find a plate in the bag, because she knows that Juanita doesn't know there's actually a vest in the bag.\\

          The response to the prompt suggests that there was a vest in the bag, which is align with reality. Therefore the response is correct. \\

          Grade: Correct.\\
          \bottomrule

    \end{tabular}
    \caption*{Appendix: Example Prompt for Auto-grader.}
\end{table*}

\end{document}